\documentclass[conference]{IEEEtran}
\IEEEoverridecommandlockouts
\usepackage{cite}
\usepackage{amsmath,amssymb,amsfonts}
\usepackage{mathtools}
\usepackage{algorithmic}
\usepackage{graphicx}
\usepackage{textcomp}
\usepackage{xcolor}
\usepackage{graphicx}
\usepackage{subfig}
\def\BibTeX{{\rm B\kern-.05em{\sc i\kern-.025em b}\kern-.08em
    T\kern-.1667em\lower.7ex\hbox{E}\kern-.125emX}}
\begin{document}

\title{Multi-scale Attention U-Net (MsAUNet): A Modified U-Net Architecture for Scene Segmentation}

\author{\IEEEauthorblockN{Soham Chattopadhyay*\thanks{*Corresponding author}}
\IEEEauthorblockA{\textit{Department of Electrical Engineering} \\
\textit{Jadavpur University}\\
Kolkata, India \\
chattopadhyaysoham99@gmail.com}
\and

\IEEEauthorblockN{Hritam Basak}
\IEEEauthorblockA{\textit{Department of Electrical Engineering} \\
\textit{Jadavpur University}\\
Kolkata, India \\
hritambasak48@gmail.com}

}

\maketitle

\begin{abstract}
Despite the growing success of Convolution neural networks (CNN) in the recent past in the task of scene segmentation, the standard models lack some of the important features that might result in sub-optimal segmentation outputs. The widely used encoder-decoder architecture extracts and uses several redundant and low-level features at different steps and different scales. Also, these networks fail to map the long-range dependencies of local features, which results in discriminative feature maps corresponding to each semantic class in the resulting segmented image. In this paper, we propose a novel multi-scale attention network for scene segmentation purposes by using the rich contextual information from an image. Different from the original UNet architecture we have used attention gates which take the features from the encoder and the output of the pyramid pool as input and produced out-put is further concatenated with the up-sampled output of the previous pyramid-pool layer and mapped to the next subsequent layer. This network can map local features with their global counterparts with improved accuracy and emphasize on discriminative image regions by focusing on relevant local features only. We also propose a compound loss function by optimizing the IoU loss and fusing Dice Loss and Weighted Cross-entropy loss with it to achieve an optimal solution at a faster convergence rate. We have evaluated our model on two standard datasets named PascalVOC2012 and ADE20k and was able to achieve mean IoU of \textbf{79.88\%} and \textbf{44.88\%} on the two datasets respectively, and compared our result with the widely known models to prove the superiority of our model over them. 
\end{abstract}



\section{Introduction}
To segment objects from different scene images has been a fundamentally challenging task for years and has been of great interest to scientists in the last two decades. This includes the detection and segmentation of regions that are correlated to semantics consisting of discrete objects like humans, animals, trees, and abstract regions like the sky, roads, and mountains from different scene images. Scene segmentation has different applications in the fields of image enhancement, robotics, self-driving cars, medical diagnosis, and many more. With ever-increasing applications of this specific task, the importance of accurate and pixel-level semantic segmentation, that allows improvements in image understandings, has been increasing. However, segmenting objects from large datasets is an intimidating task and calls for large time and resources and often provides erroneous results due to inter and intra-observer differences. Therefore the importance of computer-based semantic segmentation from complex images with finer details has increased drastically. Although there lies several challenges to tackle this issue. The change in gradients might not be conspicuous for boundary regions in some images resulting in inaccurate results which may also be resulting from different occlusions and intermittent illumination. The difference in shape, color, and size of similar objects may also be impertinent to accomplish the task of scene segmentation.
\begin{figure*}

\centerline{\includegraphics[width=2\columnwidth]{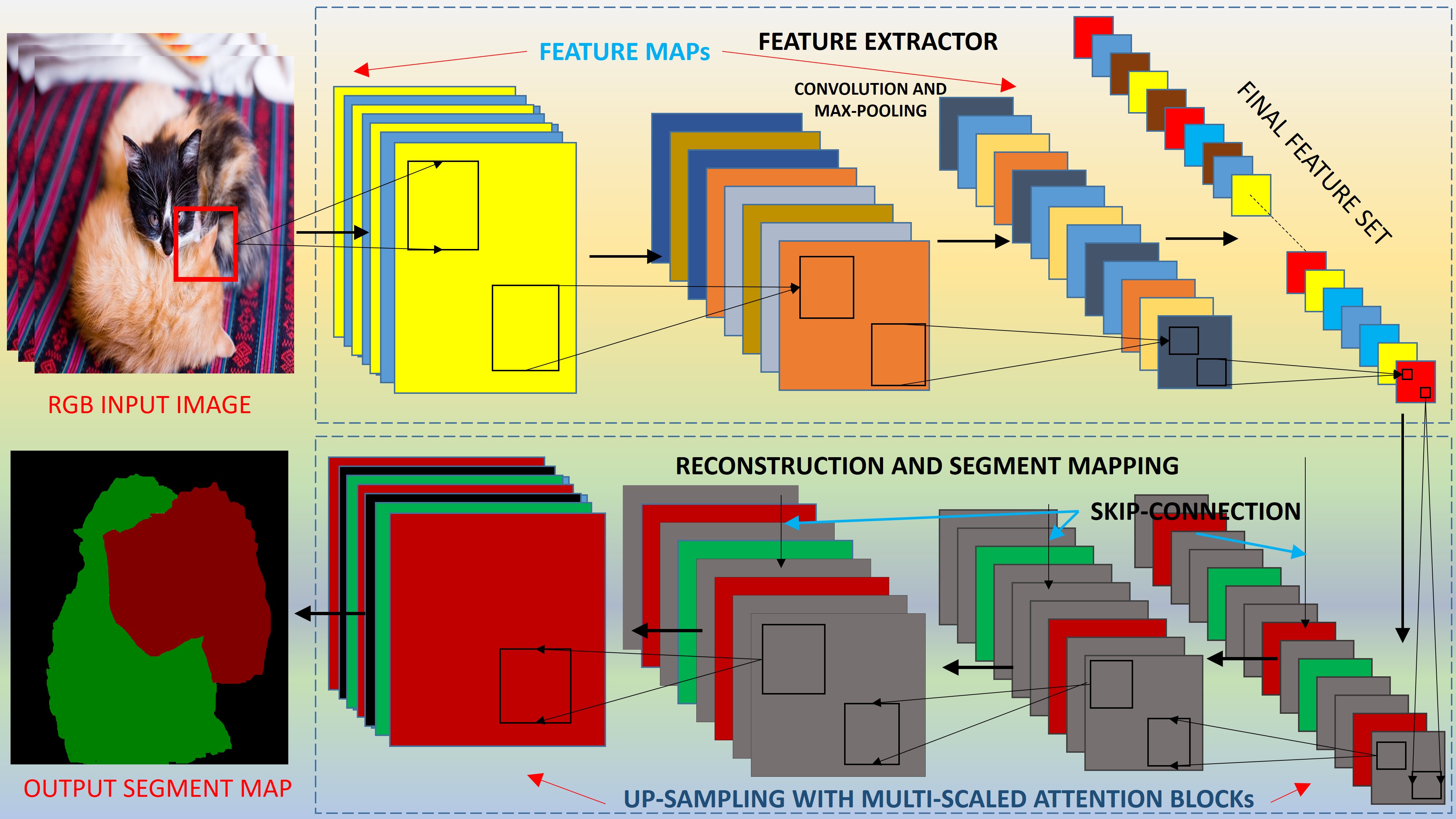}}
\centering
\caption{Workflow diagram of our proposed method}
\label{fig1}
\end{figure*}

Recently, Convolution neural networks (CNN) and fully Convolution neural networks (FCN) have been very useful to address the above problem because of its ability to extract powerful and important features for discriminative segmentation purposes.  Most of these methods have achieved state-of-the-art results in different sub-domains of image segmentation including biomedical imaging [1], image reconstruction [2], scene segmentation [3], video captioning [4], and many more. Hence, recent research works in image segmentation has shown a significant hike in the usage of CNN and encoder-decoder architecture. These architectures are composed of multi-level feature extractor which helps in contracting an image into a collection of high-level features and later expand them to produce the semantic segmentation of the input image. However, the major drawback of this solution is the presence of multiple redundant and low-level features that are extracted at different layers of the feed-forward network. Another way of utilizing the important features is to use the multi-scale attention module which concatenates the features extracted from different dilated convolution and pooling layers. Some of the existing methods [5, 6] have deployed large kernel size or introduced an additional encoding layer on the top of another which has been proven to efficiently collect the richer global context information with a decomposed structure. Even though, these strategies can be useful to segment bodies at different scales, the non-adaptive and analogous circumstantial dependencies on all image regions may lead to the absence of contextual dependencies for different image regions and the inability to segregate between global and local features. Hence the pivotal importance of multiple image segmentation problems lies in making efforts of grasping the object-relationship in the whole image.

Our proposed model possesses a multiscaling property which increases the robustness and effectiveness of the attention blocks. Multi scaling is embedded while keeping in mind that, less viable global feature maps extracted with filters having multiple scales, might gather redundancy in the reconstruction process. On the other hand, features with lower dimensions are comparatively more robust and contain more information about the contours and patterns present in the image. 
Multi-scaling techniques at extracting the local features replete the feature map with more prolific and relevant information regarding the global context of the image. So, multi-scaling techniques at extracting the local features replete the feature map with more prolific and relevant information regarding the global context of the image.Therefore, multi-scale attention upsampling with local features improves the performance, but in contrast, the performance is reduced if the same is applied to global features. With the aid of sample up-sampling with attention and multi-scaling, our proposed model produces comparable results with some of the state-of-the-art models.

\section{Related Works}
As mentioned above, Fully Convolution Networks (FCN) have been widely used for semantic segmentation and contextual aggregation has been enriched by the development of several CNN based models. The pioneers are the DeepLabV2 and DeepLabV3 [7, 8] which used atrous spatial pooling, accompanied by parallel differential dilated convolution for contextual feature enhancement. Another widespread method as mentioned above is the traditional encoder-decoder architecture. It uses a subtle mixture of different high level and low-level semantic features and combines them to estimate the relative object position [11, 12, 13]. In Fully Convolution Networks (FCN), the fully connected layers of generalized CNN are replaced by Convolution layers to achieve dense pixel predictions. The feature maps are upsampled in one or multiple upsampling layers and numbers of skip connections are also added to accomplish the improved segmentation predictions. Many extensions of these base-nets have been developed consequently [14, 15, 16, 17]. 

Attention modules improve the long-range dependencies by emphasizing on local features and eliminating the irrelevant and redundant features. Hence these have been successfully implicated in numerous computer vision problems such as image captioning, visual question answering, image enhancement, segmentation, etc. Further modifications have been made in this approach that resulted in the self-attention module [18, 19, 20] which drew the attention of scientists recently because of its ability to compute long-range dependencies with higher accuracy, keeping the computational cost and complexity small. Point-wise Spatial Attention Network (PSANet) [21] and dual attention network [22] implemented the self-attention module first for modeling the dependency of the local features and their corresponding global dependencies. The PSANet connected the local features with all others and performed robust and dynamic aggregation of contextual information through self-attention blocks. Recent findings have also proved that a single attention block might not be always sufficient enough and the output may still contain irrelevant noise from the other parts of the image, resulting in sub-optimal results [23]. To solve this problem, [23, 24] suggested multiple attention module that refines the local features by removing the undesired noises and focusing on the relevant area in the image. 

\section{Methodology}

\subsection{Network Architecture }
Our novel architecture is inspired from the attention UNet, which uses attention to give importance to a certain region out of the entire image. In general a Convolution neural network consists of some different Convolution layers and pooling layers in cascaded fashion, one on the top of other. Each Convolution layer learns different features with progression of iterations during training. Each neurone is directed related with the neurons of the previous layers via learnable weights. Each layer learns discriminative features from the image and pass the information map to the next layer. This is called forward propagation. Two consecutive layers are connected by non-linear activation functions to learn non-linear features. Then comes the pooling layer. Mainly two types of pooling layers are used in convolution layers, Max-pooling and Average-pooling. In our network we have used max-pooling to process the features while propagating from a layer to another layer. Max-pooling works as it takes the maximum value of the feature mapping over a small region of a whole image. Therefor the space invariant features are learnt by the Convolution layers and strengthen by the processing by pooling layers.

It is found by recent researchers that using static scaled filter banks can bound the space invariant feature learning capabilities of the CNN. To overcome this problem multi-scaling filter-banks are incorporated in CNN architectures. Convolution sampling at multiple scales aid CNN to learn more robust and space invariant patterns in each layer of it. In order to utilize this scope we have incorporated multi-scaled convolution layers containing  2$\times$2,4$\times$4 and 6$\times$6  convolution kernels in attention blocks while up-sampling.

Our Multi-scaled attention UNet is a simple structure, consisting of an encoder and a decoder. In our case we have used pre-trained DenseNet169 as the backbone of the architecture and used multi-scaled attention specifically to extract more viable lower dimensional features from the encoder. Particularly in computer vision tasks significant patterns of the objects should be learnt by the model in order to improve the performance of the model. In our task we have focused to learn the local features with greater importance than global features because it is discussed earlier that global features are very much sensitive towards distortions and deviations so multi-scaled global features with attention often get irrelevant redundant features and this phenomena results in ill-performance of the model. So we have omitted multi-scaling while extracting global features.

As mentioned earlier our proposed model has DenseNet169 as the encoder and 5 up-sampling block with attention as the decoder, among which 3 blocks have multi-scaled attention up-sampling blocks. These blocks mainly extract local features from the encoder architecture at multiple scales. The procedure of the multi-scaled attention up-sampling can be expressed by following:
\begin{equation}
x\textsubscript{Attention}=Attention(x,y\textsuperscript{L})
\end{equation}
\begin{equation}
x\textsubscript{1}\textsuperscript{UP}=C\textsubscript{up}(x,Kernel\: Size=2X2)    
\end{equation}
\begin{equation}
x\textsubscript{1}=Concatenate(BilinearInterpolation(x\textsubscript{1}\textsuperscript{UP},y\textsuperscript{L}),x\textsubscript{Attention} )   
\end{equation}

\begin{equation}
x\textsubscript{2}\textsuperscript{UP}=C\textsubscript{up}(x,Kernel\: Size=4X4)    
\end{equation}
\begin{equation}
x\textsubscript{2}=Concatenate(BilinearInterpolation(x\textsubscript{2}\textsuperscript{UP},y\textsuperscript{L}),x\textsubscript{Attention} )   
\end{equation}

\begin{equation}
x\textsubscript{3}\textsuperscript{UP}=C\textsubscript{up}(x,Kernel\: Size=6X6)    
\end{equation}
\begin{equation}
x\textsubscript{3}=Concatenate(BilinearInterpolation(x\textsubscript{3}\textsuperscript{UP},y\textsuperscript{L}),x\textsubscript{Attention} )   
\end{equation}

\begin{equation}
    x\textsubscript{1}'=bilinearInterpolation(x\textsubscript{1},x\textsubscript{2})
\end{equation}

\begin{equation}
    x\textsubscript{3}'=bilinearInterpolation(x\textsubscript{3},x\textsubscript{2})
\end{equation}

\begin{equation}
    x\textsubscript{o}=C\textsubscript{f}(Concatenate(x\textsubscript{1}',x\textsubscript{2},x\textsubscript{3}'),Kernel\: Size=1\times1)
\end{equation}
where C\textsubscript{UP} and C\textsubscript{f} are linear convolution operations, first one has different scaling properties with fixed input and output channels and the second one has input is equal to three times the output and having a kernel size of 1. Technically which means all versions of C\textsubscript{up}$\in$ $\mathbb{R}$\textsuperscript{$Ch\textsubscript{in}\times Ch\textsubscript{out}$} and C\textsubscript{f}$\in$ $\mathbb{R}$\textsuperscript{$3*Ch\textsubscript{out}\times Ch \textsubscript{out}$}. Aforementioned sequential mathematical expressions of the multi-scaled up-sampling attention block is described pictorially in Fig. 2(B)

\begin{figure*}
\begin{center}
  \includegraphics[height=105mm]{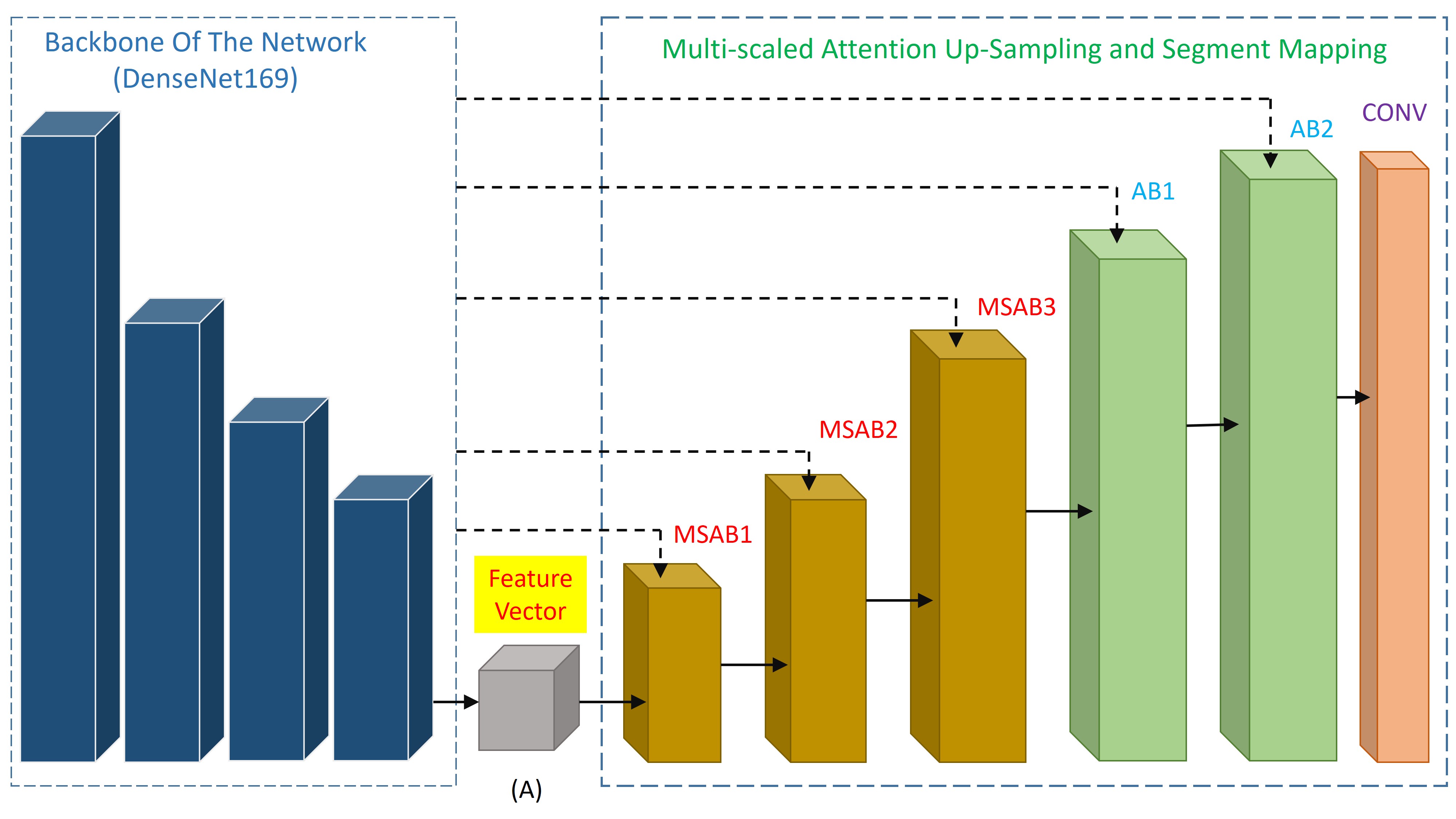}

\label{fig:2A} 

\end{center}

\begin{center}
  \includegraphics[height=105mm]{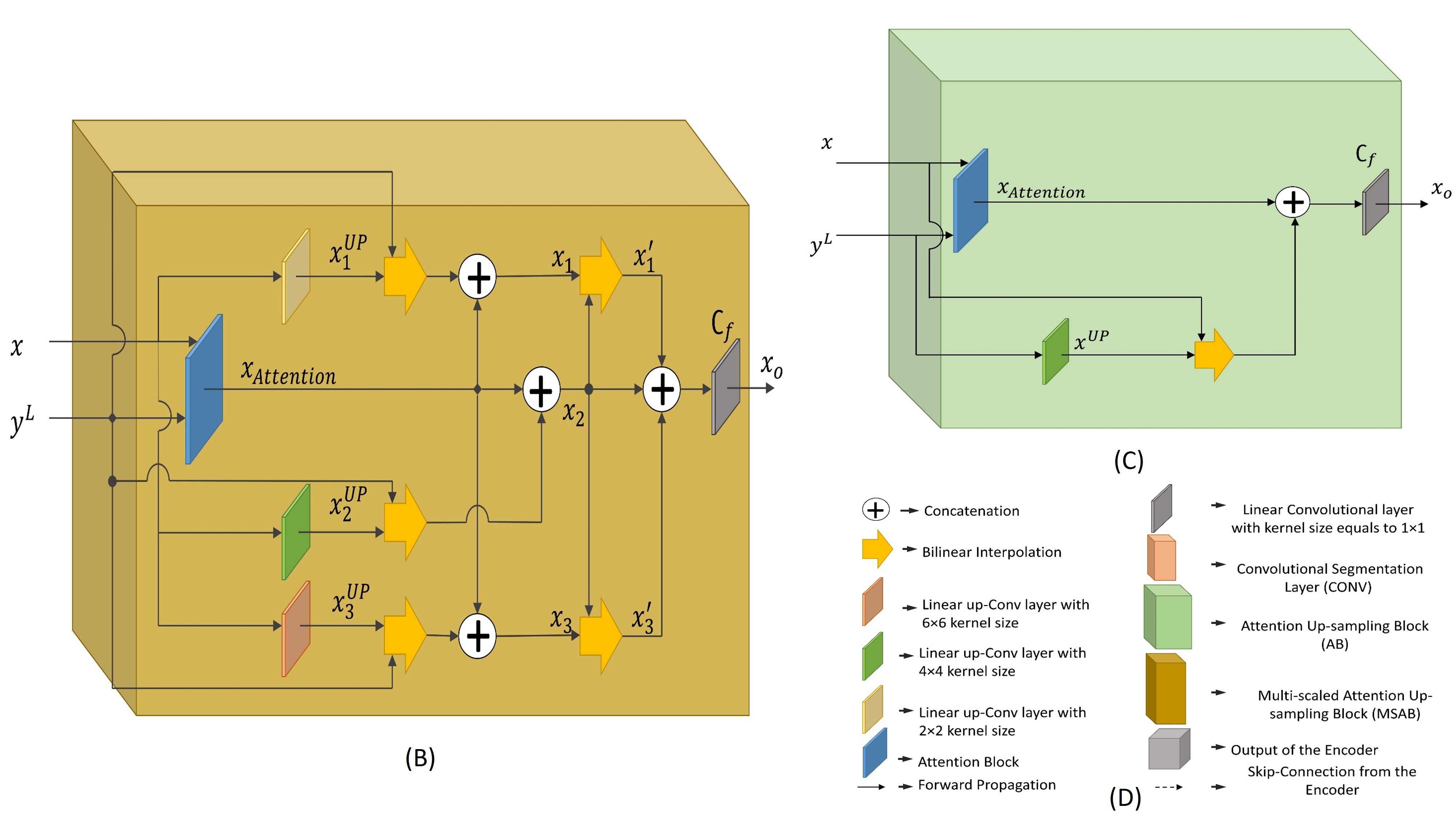}
\caption{(A)Proposed MsAUNet, (B)Multi-scaled attention up-sampling block, (C) Single-scaled attention up-sampling block (D)Specifications of different operations }
\label{fig:2B} 

\end{center}
\end{figure*}

In our network we have discussed that we embedded five such up-sampling layers first three have multi-scaled attention in it and later two layers have no multi-scaling but attention. These transpose convolution layers consist of only a single 4$\times$4 Convolution kernel. In addition to these five up-sampling layers there is also a single down convolution layer having kernel size equal to 1 and which maps to the number of classes that the ground truth images have. Each transpose convolution block is followed by Batch normalisation layer and is connected by the layer pyramid pooling layer with a nonlinear activation function Leaky ReLU to learn nonlinear orientations or patterns of the image for segmentation mappings. For skip connection we have extracted features from different layers of DenseNet169 which are namely: \textbf{ None, 'relu0', 'denseblock1', 'denseblock2', 'denseblock3'}. And the main feature extracted from the output layer of the DenseNet169 that is \textbf{'denseblock4'}. The output of the DendeNet169 has 2208 channels with 7 $\times$ 7 output height and width of the feature map. The multi-scaled attention up-sampling blocks(MSAB) as well as the attention up-sampling blocks(AB) each up-samples with a scale of 2. The detailed representation of MSAB layers and AB layers is given in Fig. 2(B) and Fig. 2(C) respectively. The output of the DenseNet169 is mapped via first multi-scaled attention up-sampling bock to 256 channels and the dimension is increased by 2 folds. Similarly 256 feature map is mapped to 128 and 128 to 64 by similar multi-scaled attention blocks for up-sampling. These three blocks take the output of 'denseblock1', 'denseblock2', 'denseblock3' as skip-connections in reverse order from the last respectively. In the later part of the architecture two normal attention up-sampling blocks maps 64 channels to 32 and 32 to 16. These two blocks take the skip-connections as the output of the rest two blocks in reverse order. So the linear Convolution operations discussed so far produce an output of shape  16 $\times$ 224 $\times$224 which is further fed to a Convolution layer of kernel size is equal to 1 $\times$ 1  which classifies each pixels to a certain class. Pictorial representation of proposed Multi-scaled Attention U-Net(MsAUNet) is given by Fig. 2(A).

\subsection{Attention Block}
In our proposed work we have used attention mapping [25] in multi-scaled fashion at the pyramid pooling layers of our network. Different from the original UNet architecture we have used attention gates which take the features from the encoder and the output of the pyramid pool as input and produced output is further concatenated with the up-sampled output of the previous pyramid-pool layer and mapped to next subsequent layer. The schematic diagram of attention block is given by Fig. 3.

Attention block produces a constant value, so called Attention Coefficient $\beta_m$ which lies between 0 and 1 identifies significant region out of the whole image itself. Attention block produces the output which is the element-wise multiplication of the attention coefficient and the output of the prior defined layer shown by the following equation.
\begin{equation}
    y\textsubscript{m}\textsuperscript{L\textsubscript{a}} = y\textsubscript{m}\textsuperscript{L} \cdot \beta _m
\end{equation}
Where y\textsubscript{m}\textsuperscript{L\textsubscript{a}} is the output of attention block with attention map in it and have the dimensions same as that of the dimensions of input feature map y\textsubscript{m}\textsuperscript{L} having for L\textsuperscript{th} layer and m\textsuperscript{th} pixel. It is quite evident from above equation that attention block generates an Importance value (attention coefficient) for every pixels and specifies the significance of a particular region by multiplying the pixels with their corresponding attention coefficient. If a region has $\beta\textsubscript{m}$ close to 1 then the region is of higher importance compare to the region having importance value near to 0, which gets relatively lower importance. So the attention blocks learns to focus at certain region of an entire image. If the segmentation task has multiple classes then the network learns multi-dimensional importance values, this idea is proposed by [26].
\begin{figure}
\begin{center}
  \includegraphics[width=\columnwidth,height=60mm]{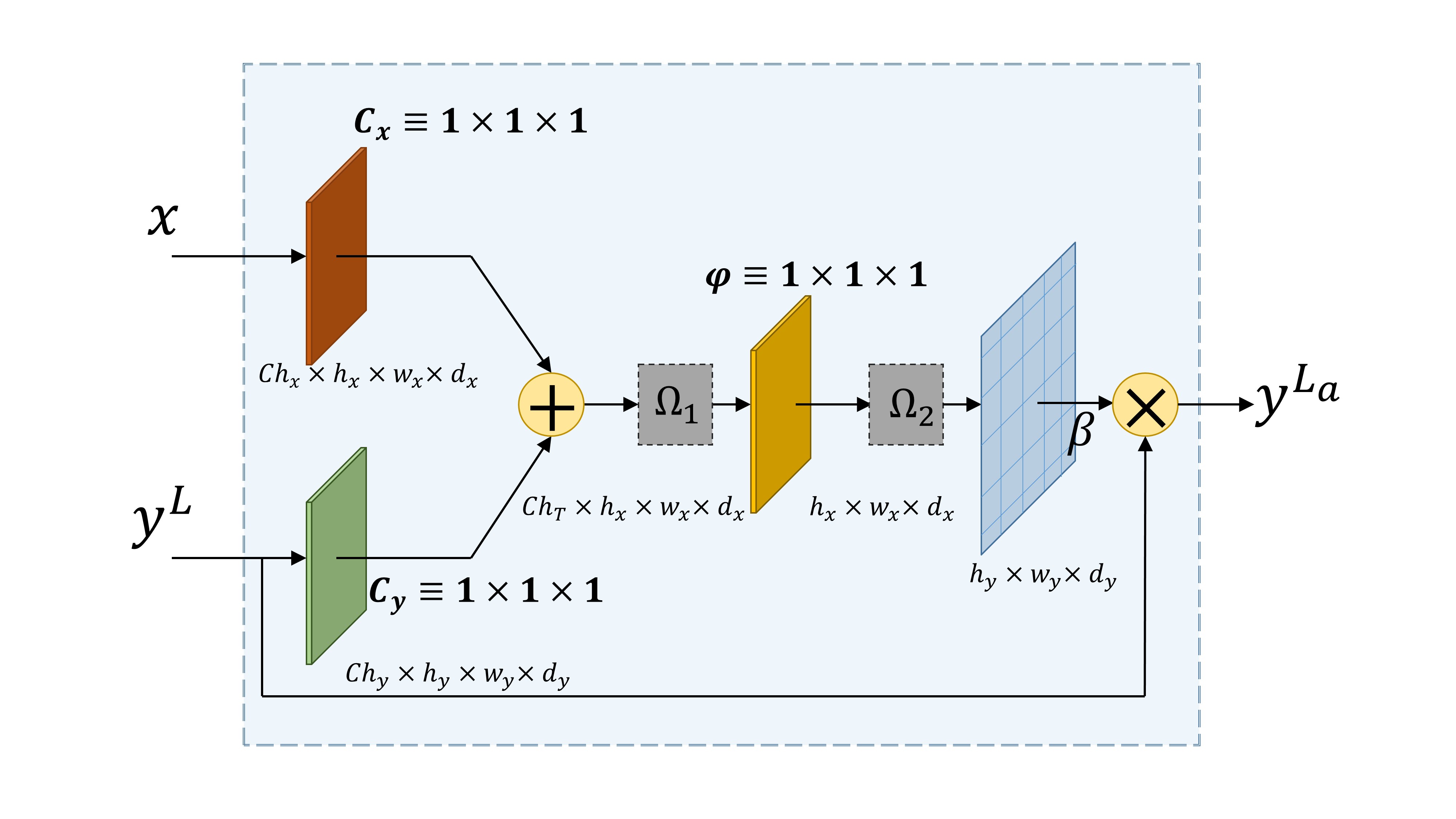}
\caption{Attention Block Function}
\label{fig:3} 

\end{center}
\end{figure}

The attention block takes the feature vector of a specific layer of the encoder y\textsuperscript{L} through skip-connection and the input feature map of the previous convolution block x of the architecture as inputs, and they are passed through two different convolution layers, each having kernel size of 1$\times$1 to bring two inputs into same number of channels without changing their dimensions. Thereafter an up-sampling operation is performed on the input feature map to have same size as that of  x and they are concatenated and passed through ReLU activation function. Now the output of the ReLU is passed through another 1 $\times$ 1 convolution followed by a Sigmoid activation function to get a flatten output having a score between 0 and 1 indicating how much importance to be given to a particular region of the image. Then the skip input is multiplied with the attention map producing the final output of the attention block. From Fig. 3 we can easily express the idea of attention with following mathematical terms:
\begin{equation}
    a_m=C_x^T x_m+C_y^T y_m^L+b_x,
\end{equation}
\begin{equation}
    b^L=\phi ^T (\Omega _1 (a_m ))+b_\phi
\end{equation}

Using above two equations we can simply express that, 
\begin{equation}
\beta  _m^L=\Omega_2 (b^L (y_m^L,x_m,\varepsilon_{\rm att} )) \end{equation}
Where $\beta_m^L$ and $\Omega _1 ()$ are respectively the ReLU and Sigmoid activation functions. $\epsilon_{att}$ contains the linear transformation parameters $C_x \in \mathbb{R} ^{Ch_x \times Ch_T}$ , $C_y \in \mathbb{R} ^{Ch_y \times Ch_T}$ and $\phi  \in \mathbb{R} ^{Ch_T \times 1}$ the bias terms  $b_\phi \in \mathbb{R} , b_x \in \mathbb{R} ^ {Ch_T}$.
\subsection{Loss Function}
We have formed a compound loss as the weighted sum of some popularly used segmentation loss functions as our final loss function. The loss function is given as follows: 
\begin{equation}
    L_F=L_{\rm iou}+0.01*L_{\rm Dice}+0.8L_{\rm WCE}
\end{equation}

\subsubsection{Weighted Cross Entropy Loss}

WCE loss is an extended version of Cross Entropy loss where different weights are assigned to different classes such that un-presented classes are allocated with larger weights. In semantic segmentation problem WCE loss is one of the most commonly used loss function [27, 28, 29], where for a given pixel j in an image I\textsubscript{k} the loss calculation would be as follows
\begin{equation}
L\textsubscript{WCE}=-\sum\limits_{j}\sum\limits_{n_1}^n\theta\textsubscript{k}\textsuperscript{j}G\textsubscript{k}^{(j,n_1)} log(P\textsubscript{k}^{(j,n_1)})
\end{equation}
where $\theta\textsubscript{k}\textsuperscript{j},\:G\textsubscript{k}^{(j,n_1)}$ and $P\textsubscript{k}^{(j,n_1)}$ are respectively the weights allocated to pixel j, the ground truth value which is one for the pixel which belongs to the class $n_1$.Here n is the number of classes in the segmentation task. Inspired from [39], the pixels near the boundary regions are difficult to identify so we have put more weights to those pixels specifically. For pixel j and image $I_k$ the calculated weight is
\begin{equation}
    \theta_k^{(j)}=1+\epsilon_1\ddot{I}\left(|\nabla G_k^j|>0\right)+\epsilon_2\ddot{I}\left(G_k^j=C\right)
\end{equation}
In the above equation $\epsilon_1$ and $\epsilon_2$ are constants and $\ddot{I}()$ is an indicator function which gives one for true inner statements and gives zero for other values. $C$ is the underground classes and $\nabla$ is the divergent. 

\subsubsection{Dice Loss}
This special region based loss is maximally used to reduce the error caused by overlapping of the prediction and the ground truth labels, proposed by [40]. To use Dice Loss we have to use Softmax activation function to normalize all pixel values in between 0 and 1. The specialty of dice loss is it can deal very well with class imbalance problems. Dice loss can be expressed by the following equation:
\begin{equation}
    L_{Dice}=\sum\limits_{n\in N}1-\frac{2\sum\limits_{j\in I_k}G_k^{(j,n)}A_{sc}\left(P_k^{(j,n)}\right)+\alpha}{\sum\limits_{j\in I_k}\left(\left(G_k^{j,n}\right)^2+A_{sc}\left(P_k^{j,n}\right)^2\right)+\alpha}
\end{equation}
For pixel j in image $I_k$ with estimated probability $P_k^{j,n}$ and ground truth label $G_k^{j,n}$. $A_{sc}$ is the softmax activation function. 

\subsubsection{IoU Loss and its optimization}
IoU loss is similar to Dice loss, which also directly optimize the segmentation mismatch error. As we know the output of the network is the probability of a pixel being in a certain class of region. Paper [41] proposes a good approximation of IoU loss for segmentation problems that we have implemented in our proposed model.If $Y_0$ is the output of the model containing pixel probabilities of the of each pixels belonging to the pixel vector $P={1,2,3,.....,n}$. and if the ground truth values of each pixel from above pixel vector is $Y_G\in \{0,1\}^P$ where 0 represents the background class and 1 represents the object then the calculation of IoU is given by:
\begin{equation}
IoU= \frac {I(Y_O)}{U(Y_O)}
\end{equation}
Where the approximated $I(Y_0)$ is given as follows:
\begin{equation}
    I(Y_O )=\sum_{\rm p\in P}Y_O^p*Y_G^p
\end{equation}
And the approximate value of $U(Y_0)$ is given in the following expression:
\begin{equation}
I(Y_O )=\sum_{\rm p\in P}Y_O^p+Y_G^p-Y_O^p*Y_G^p 
\end{equation}
So the IoU loss can be calculated as
\begin{equation}
    L_{\rm IoU}=1-\frac{I(Y_O)}{Y_O}
\end{equation}
Where $L_{IoU}$ represents the IoU loss. Now if $\theta$ represents the parameters (weights and biases) of the proposed network and if $L_{IoU}$ is incorporated into the objective function then it can be represented as
\begin{equation}
   arg\: min_{\theta}\:L_{IoU}=1-\frac{I(Y_O)}{U(Y_O)}
\end{equation}

To achieve the optimal set of trainable parameters that is, the solution of above equation is found using stochastic gradient descent which is calculated with respect to the output of the network is given by
\begin{equation}
\frac{\partial L_{\rm IoU}}{\partial Y_O^p}=-\frac{\partial}{\partial Y_O }\left [ \frac{I(Y_O )}{U(Y_O )} \right ]
\end{equation}

Which can be further explained as 

\begin{equation}
\frac{\partial L_{\rm IoU}}{\partial Y_O^p }=\frac{-U(Y_O )*\frac{\partial I(Y_O )}{\partial Y_O^p}+I(Y_O )*\frac{\partial U(Y_O )}{\partial Y_O^p}}{U(Y_O )^2}
\end{equation}
\begin{equation}
or, \frac{\partial L_{\rm IoU}}{\partial Y_O^p }=\frac{-U(Y_O )*Y_G^p+I(Y_O )*(1-Y_G^p)}{U(Y_O )^2}
\end{equation}

The simpler version of above expression is given by as follows:

\begin{equation}
    \frac{\partial L_{\rm IoU}}{\partial Y_O^p }=\left\{\begin{array}{lr}
    -\frac{1}{U(Y_O)}\:if\:Y_G^P=1\\
    \frac{I(Y_O)}{U(Y_O)^2}\:otherwise
    \end{array}\right\}
\end{equation}

If the gradient function is calculated we can simply find the derivatives of the objective function by simple back-propagation of the gradients with respect to the parameters of the network $\theta$.

\section{Experiments}
We have trained our model with PascalVOC2012 and ADE20k dataset and fine-tuned the model with hyperparameter tuning. the details of experimentation are discussed below.To evaluate the performance we have calculated some evaluation criterions such as pixel accuracy, mIoU, frequency weighted IoU and Dice coefficient (F1 Score).
\subsection{Datasets}
\subsubsection{PascalVOC2012}PascalVOC [42] dataset is used to recognize objects from numerous object classes in photo-realistic scenes. It contains 20 different classes consisting of different indoor and outdoor scene objects. Some of the sample classes are human, different animals (dog, cow, bird, cat etc.), different vehicles (car, bus, train etc.) and different indoor objects. There are segmented images used for training and based on the annotation data, a model is trained for segmentation or classification. The dataset may also be used for action classification such as driving, walking, dancing etc. The dataset contains 10,582 train images, 1,449 validation images and 1,456 test images along with their ground truth labels.
\subsubsection{ADE20k}
ADE20k is a widely used dataset which has multipurpose application including scene segmentation, object detection etc. [43].The dataset contains images of indoor as well as outdoor scenes with their corresponding annotation images. The dataset is split in train, test, validation sets and split in proper tree hierarchy. The R and G channels of the RGB images are used to represent the object class masks and the B channel indicate the instance object class.

\subsection{Pixel Accuracy}
Pixel accuracy is simply the percentage of pixels has been classified correctly in an entire image. Relying on only Pixel Accuracy is certainly not a very good choice especially if class imbalance problem occurs in majority of the images. But in our datasets such problems don’t occur such readily so it took a good role in evaluating our model’s performance.

Suppose $p_{nm}$ be the total number of pixels classified to class m but originally belong to class n among total N number of classes and if total number of pixels in class  n is denoted by  $T_n=\sum\limits_n P_{nm}$, then pixel accuracy can be calculated as 
\begin{equation}
    Pi_A=\frac{\sum_n p_{nn}}{\sum_n T_n}
\end{equation}
\subsection{Mean IoU}
Intersection over Union also known as Jaccard Index is one of the mostly used evaluation metrics for segmentation problem. IoU is simply the ratio of the area of intersection between the predicted segmentation and ground truth label to the sum of these areas for a single class of segmentation. Mean IoU is the mean of all IoUs in a particular image having more number of classes. Mean IoU matric should be ideally 1 for perfect prediction but usually if the value is more than 0.6 it is considered to be a good prediction for object detection problems.

Following previously mentioned convention we can interpret the mathematical expression of mean IoU as given by the following equation:
\begin{equation}
    mIoU=\frac{1}{N} \left ( \frac{\sum_n p_{nn}}{T_n+\sum_m p_{mn} -p_{nn}} \right )
\end{equation}

\subsection{Frequency weighted IoU}
This evaluation matric is similar to Mean IoU with adequate weighted frequency values multiplied with corresponding pixels. The expression of fWIoU is as follows
\begin{equation}
    fIoU=(\sum_s T_s )^{-1} \left ( \frac{\sum_n T_n*p_{nn}}{T_n+ \sum_m p_{mn} -p_{nn}} \right )
\end{equation}

\subsection{Dice Coefficient}
Dice coefficient, popularly known as F1 Score is the ratio of two times the overlapping area of ground truth and predicted segmented area of a certain class to the total value of ground truth and segmented area. In an image for a certain class if $A_{GT}$, $A_{SG}$ are the areas of ground truth image and predicted segmented image for a certain class in the image then the Dice Coefficient is calculated as 
\begin{equation}
    D_C=2*\frac{A_GT\cap A_SG}{A_GT\cup A_SG}
\end{equation}

\subsection{Comparison with Different Backbone Architectures}
To optimize the performance of our model we have performed several experiments, one of the most important task is to fix the backbone. To do so we have kept the decoder part intact and keep changing the backbone model of our architecture. We have done exhaustive experiments considering different backbone models, some of them such as \textbf{AlexNet, ResNet18, ResNet50, VGG11, VGG13, GoogLeNet} and so on has not performed so good. On the other hand \textbf{ ResNet101, ResNet152, VGG19 and DenseNet169} as the backbone architecture has performed very well. We can simply interpret the prime criterion of demarcation between previously mentioned nets and the later ones is the depth of each network. It is quite evident that in segmentation problems the neural net is to learn the features which are basically complex patterns out of a whole image. So more is the depth of the network, more it learns important patterns and produces more robust local features, containing relevant information about the complex patterns of the image which are further carried on to decoder through skip-connections and multi-scaled attention blocks. Therefore if the local features are pretty good themselves then the decoder also consists of very prolific features which produces better output.

Along with aforementioned performance metrics we also have given the converging train and validation loss plots of our model for different backbone architectures for comparison purpose.

\begin{figure*}

\centerline{\includegraphics[width=1.6\columnwidth]{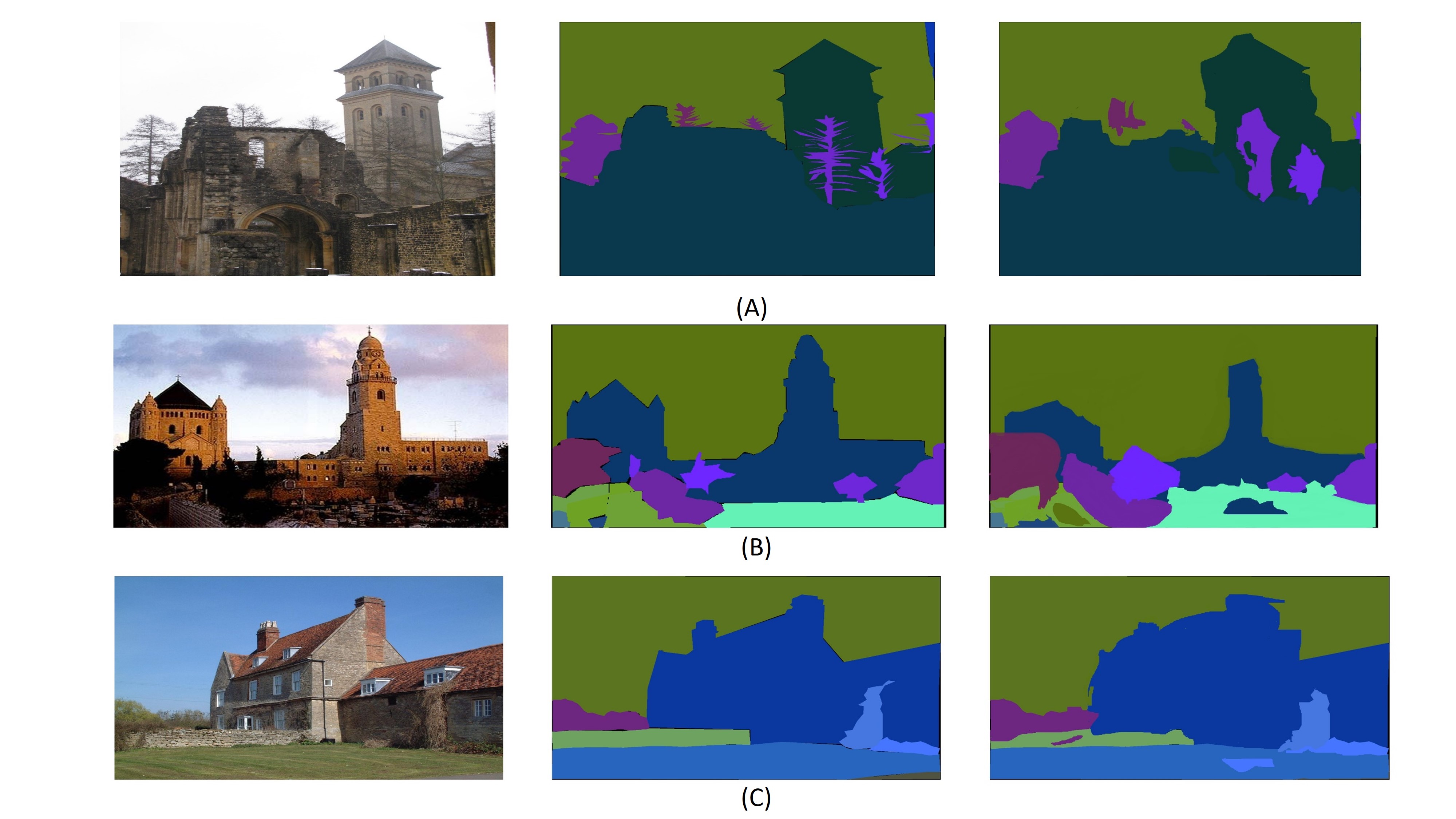}}
\centering
\caption{Segmentation of our proposed method on ADE20k dataset: the first column represent the original input image, the second column represent the corresponding ground truth and the last column is the output image.}
\label{fig4}
\end{figure*}
\subsection{Hyper-parameter Tuning}
Along with aforementioned experiments to optimize our architecture we also have performed some experiments, explaining the converging criterions to reach the optimum performance. For that purpose we have fixed the backbone with DenseNet169 (It gives maximum mIoU as discussed in the following Results and Discussion section) and changed the learning rate, optimizer and mainly the hybrid loss, which is the weighted sum of different region-based and distribution-based loss. In case of optimizer, we have compared the performance of the model with three standard optimizers, such as SGD, RMSPROP and ADAM with changing learning rates of 1e-3, 1e-4, 1e-5 and 1e-6 for both datasets PascalVOC2012 and ADE20K. The comparative performance is shown by a bar diagram and discussed in the following section. In addition, the weighted parameters of hybrid loss function is also optimized. After certain experimentation, we finalize the loss function to be like it is mentioned above. The optimizer and learning rates carry significant values in the performance improving of the model. So, we become very careful while choosing appropriate optimizer and learning rates.

We have performed aforementioned experiments for both PascalVOC2012 and ADE20K datasets and the appropriate tables and figures mentioning the results are given in the Section 5, Results and Discussion. 
\section{Results and Discussions}
We have evaluated our model on popularly used PascalVOC2012 dataset and ADE20K. As we have discussed earlier in section 6.5 that we have used different backbone architectures and checked the performance of the model in each case to achieve the optimum performance with our proposed framework. Table 1 and Table 2 represent comparative studies of our proposed model with different deep Convolution networks as encoders on PascalVOC2012 and ADE20K datasets. Both Table 1  and Table 2 show that DenseNet169 performs the best as a feature extractor which brings about \textbf{95.96\%} and \textbf{81.41\%} pixel accuracies with \textbf{79.88\%} and \textbf{44.59\%} mean IoU on PascalVOC2012 and ADE20K datasets respectively. In addition VGG19 and ResNet152 also give promising results with 76.17\% and 77.19\% mIoUs in PascalVOC2012 and 43.97\% and 43.20\% mIoUs in ADE20K datasets respectively.
\begin{table*}
\caption{Comparison of performances with different backbone architectures of MsAUNet on PascalVOC2012 dataset}
\resizebox{\textwidth}{!}{\begin{tabular}{|c|c|c|c|c|}
\hline
Backbone Architecture & Pixel Accuracy (\%) & Mean IoU (\%) & Frequency Weighted   IoU (\%) & Dice Coefficient (\%) \\ \hline
ResNet101             & 95.46\%             & 71.32\%       & 77.23\%                       & 74.36\%               \\ \hline
ResNet152             & 93.21\%             & 77.19\%       & 80.12\%                       & 78.96\%               \\ \hline
VGG19                 & 94.00\%             & 76.17\%       & 79.33\%                       & 77.64\%               \\ \hline
ResNet18              & 88.45\%             & 63.44\%       & 67.55\%                       & 62.19\%               \\ \hline
DenseNet169           & 95.96\%             & 79.88\%       & 80.79\%                       & 79.01\%               \\ \hline
\end{tabular}}

\caption{Comparison of performances with different backbone architectures of MsAUNet on on ADE20K}
\resizebox{\textwidth}{!}{\begin{tabular}{|l|l|l|l|l|}
\hline
Backbone Architecture & Pixel Accuracy (\%) & Mean IoU (\%) & Frequency Weighted   IoU (\%) & Dice Coefficient (\%) \\ \hline
ResNet101             & 79.22\%             & 41.45\%       & 46.94\%                       & 44.56\%               \\ \hline
ResNet152             & 80.19\%             & 43.20\%       & 50.35\%                       & 45.39\%               \\ \hline
VGG19                 & 78.02\%             & 42.97\%       & 45.23\%                       & 50.04\%               \\ \hline
ResNet18              & 67.13\%             & 33.24\%       & 34.67\%                       & 42.31\%               \\ \hline
DenseNet169           & 81.41\%             & 44.59\%       & 54.48\%                       & 51.01\%               \\ \hline
\end{tabular}}
\end{table*}

While dealing with deep neural networks, convergence plot of train and validation losses over iterations is a very good measure of comparison. While training our model with 1200 epochs, different backbone architectures show significantly different characteristics in terms of train-validation loss plots also. Fig. 5 and Fig. 6 shows the train-validation loss plots for PascalVOC2012 and ADE20K with aforementioned backbone encoders. It is very much intuitive from both of the figures that DenseNet169 converges the mostly, around 1000 epochs. On the other hand ResNet18 starts over-fitting just after epoch 600. ResNet101 and ResNet151 show similar characteristics for both datasets and VGG19 does not show any fixed pattern. From Table 1 , Table 2 and Fig. 5 , Fig. 6 we can conclude that DenseNet169 performs better as the feature extractor than any other deep nets.
\begin{figure*}

\centerline{\includegraphics[width=2\columnwidth]{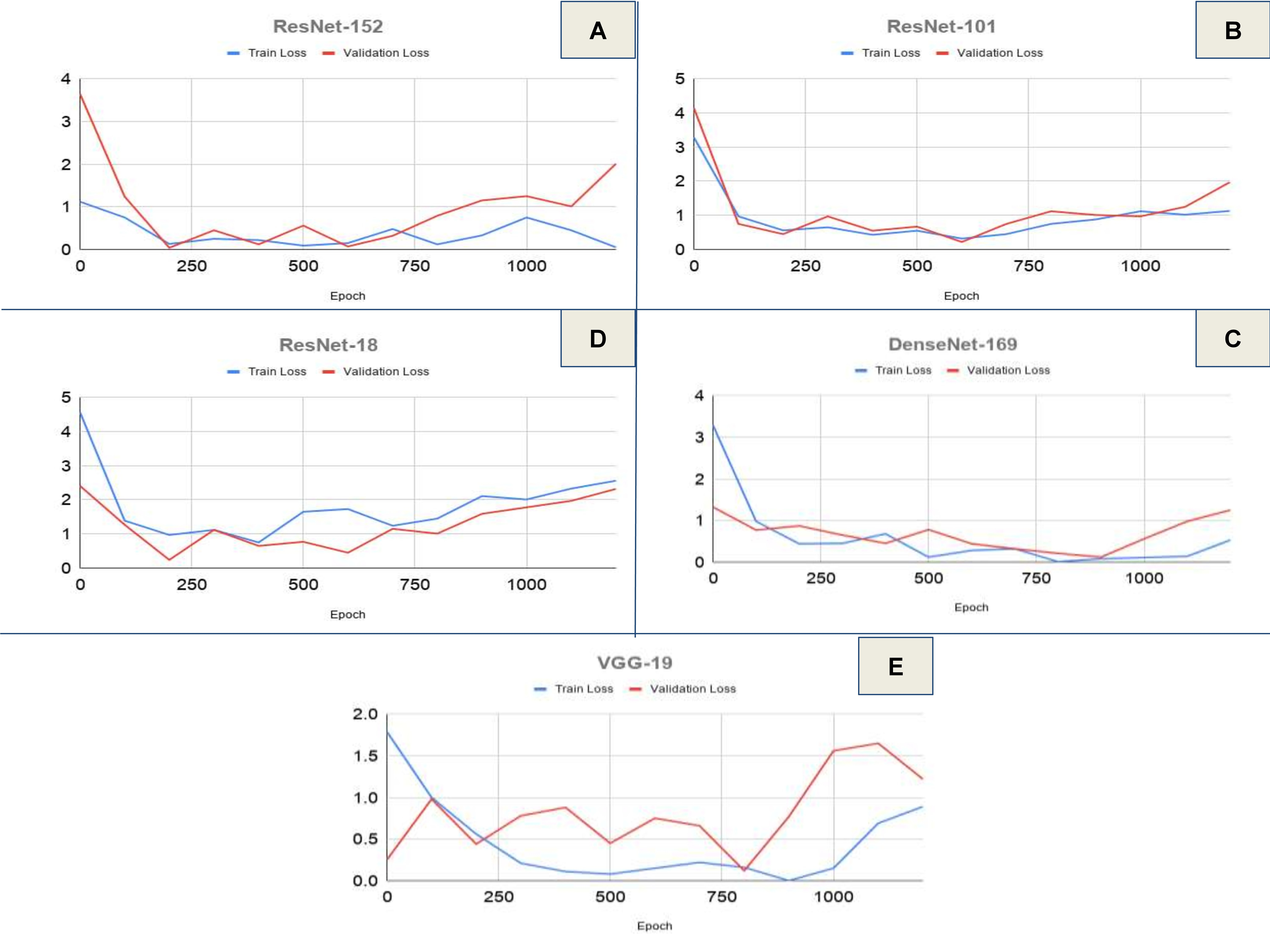}}
\centering
\caption{A comparative study of loss plots for different deep neural architectures as the backbone of MsAUNe on PascalVOC2012 dataset: (A)variation of Train and Validation loss plots with respect to epochs with ResNet152 as the backbone architecture.(B)variation of Train and Validation loss plots with respect to epochs with ResNet101 as the backbone architecture.(C)variation of Train and Validation loss plots with respect to epochs with ResNet18 as the backbone architecture.(D)variation of Train and Validation loss plots with respect to epochs with DenseNet169 as the backbone architecture.(E)variation of Train and Validation loss plots with respect to epochs with VGG19 as the backbone architecture.}
\label{fig5}
\end{figure*}
\begin{figure*}

\centerline{\includegraphics[width=2\columnwidth]{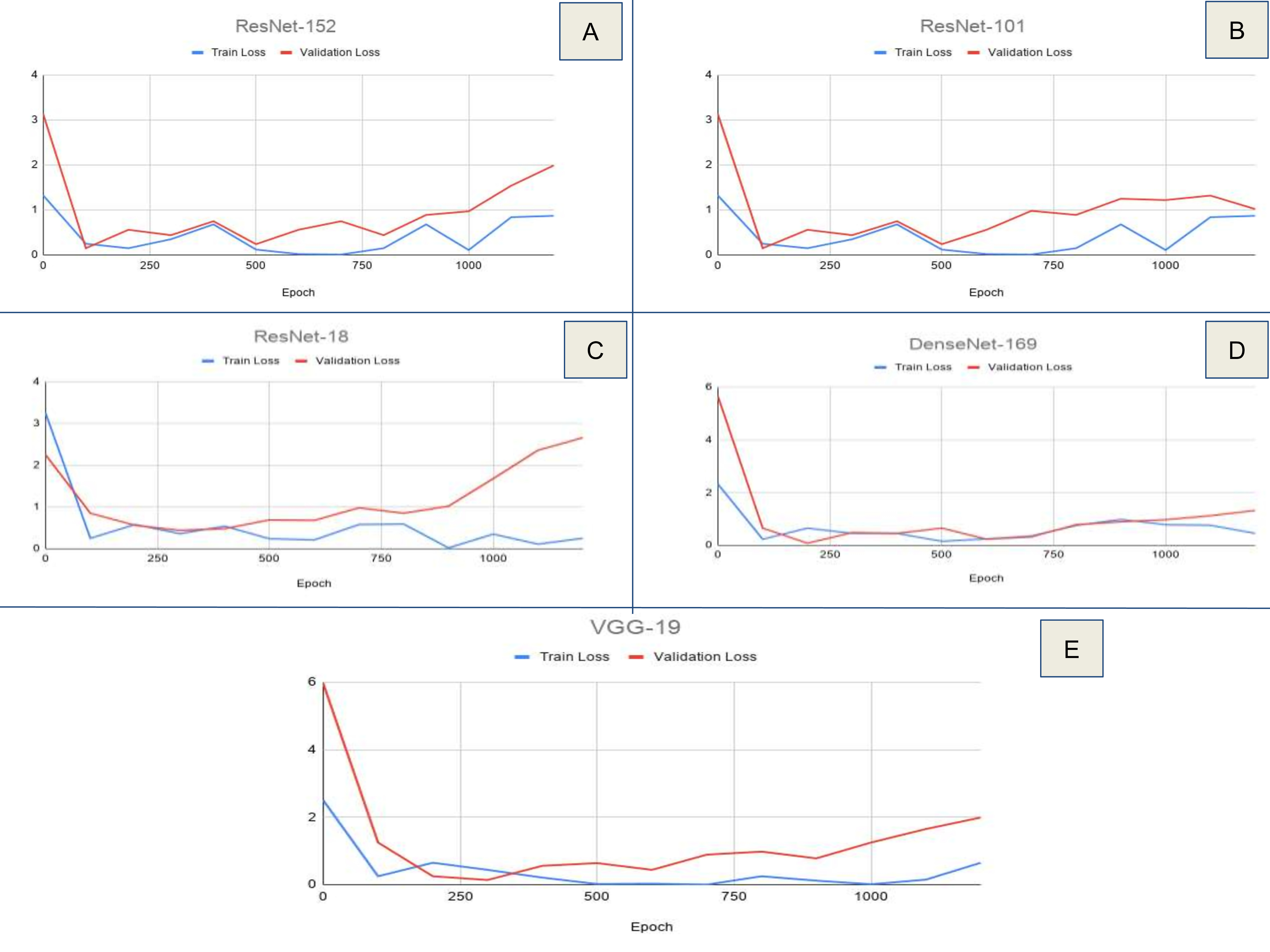}}
\centering
\caption{A comparative study of loss plots for different deep neural architectures as the backbone of MsAUNe on ADE20K dataset: (A)variation of Train and Validation loss plots with respect to epochs with ResNet152 as the backbone architecture.(B)variation of Train and Validation loss plots with respect to epochs with ResNet101 as the backbone architecture.(C)variation of Train and Validation loss plots with respect to epochs with ResNet18 as the backbone architecture.(D)variation of Train and Validation loss plots with respect to epochs with DenseNet169 as the backbone architecture.(E)variation of Train and Validation loss plots with respect to epochs with VGG19 as the backbone architecture.}
\label{fig6}
\end{figure*}
Appropriate choices of learning rate and optimizer plays vital roles in the performance of any deep neural network. Therefore to optimize our model’s performance with appropriate choices of hyper-parameters, we have done exhaustive experiments with varying learning rates and optimizers and checked how the mIoU is varying for each datasets. The experiment results are shown using bar diagrams in Fig. 7. It is seen that SGD with high learning rates performs better than the other two in PascalVOC2012 with good margin of difference. Whereas in ADE20K dataset with learning rate of 1e-3, Adam optimizer outperforms others. In PascalVOC2012 dataset, RMSPROP optimizer with learning rate of 1e-5 gives comparable results with Adam optimizer and learning rate of 1e-4. In general for our case Adam optimizer performs better than other two optimizers with learning rate of 1e-4 and 1e-5 on PascasVOC2012 and ADE20K datasets respectively.

\begin{figure}
\begin{center}
  \includegraphics[width=\columnwidth]{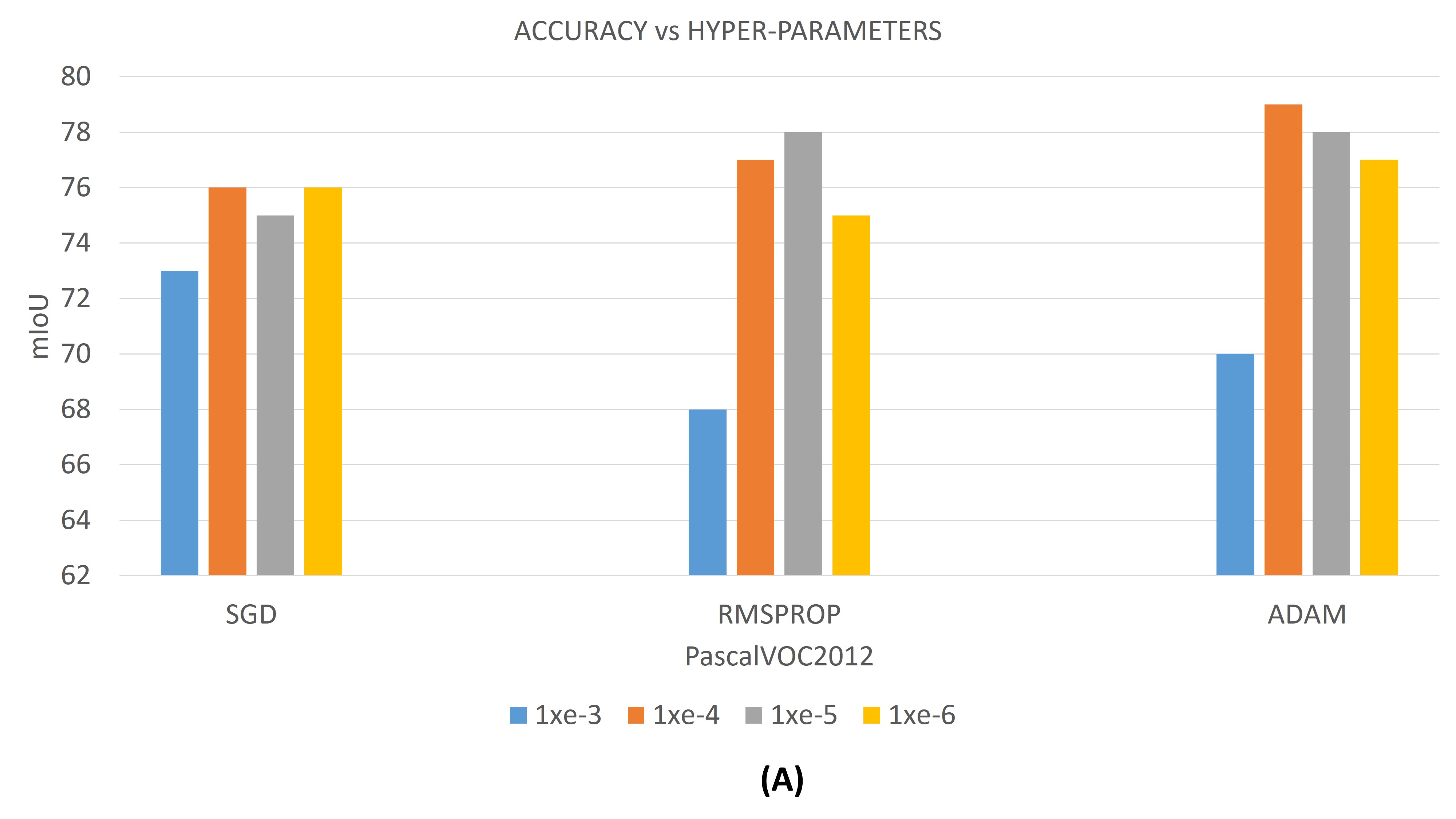}

\label{fig:1} 

\end{center}

\begin{center}
  \includegraphics[width=\columnwidth]{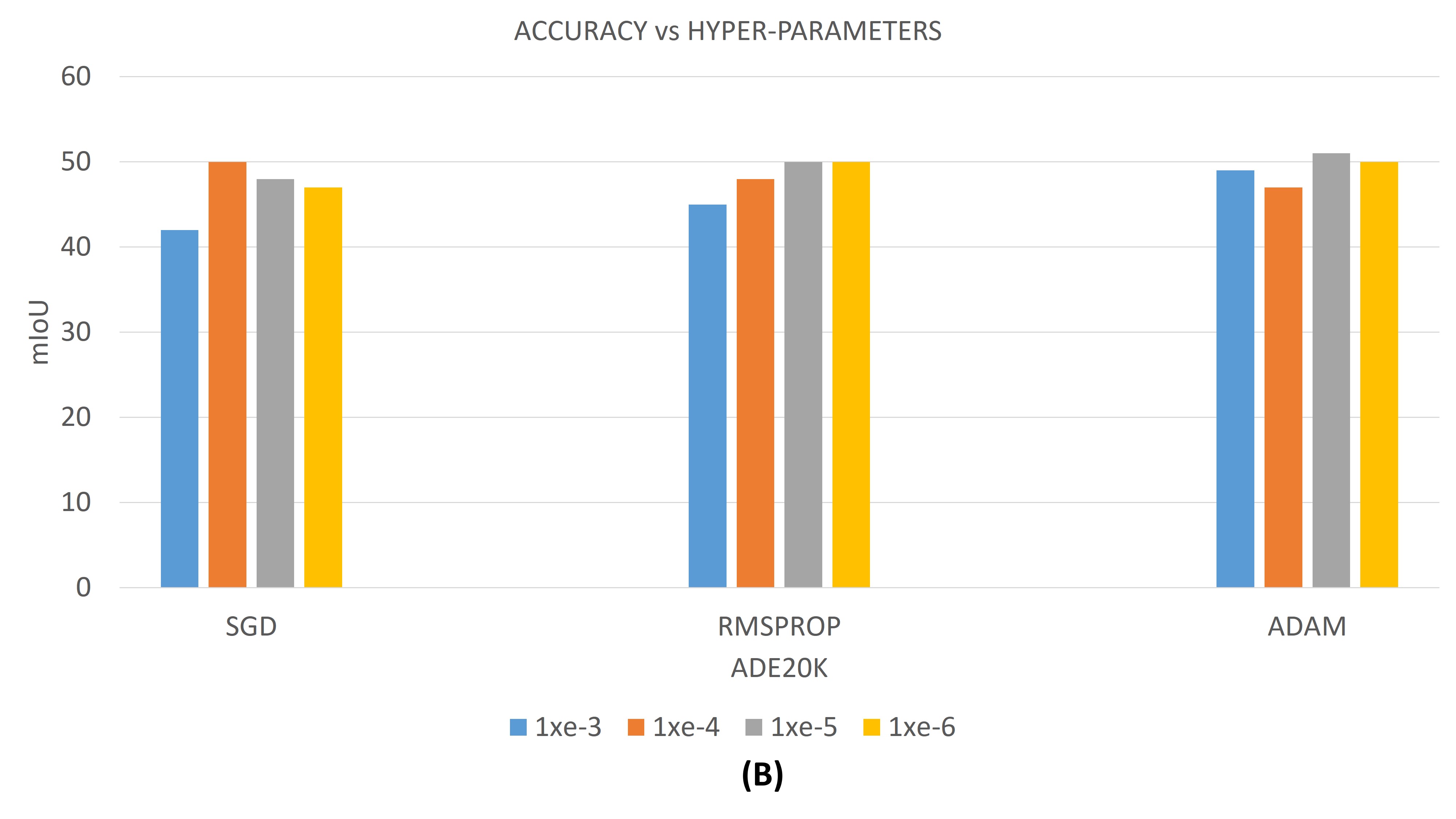}
\caption{Variation of mIoU with respect to different optimizer at different Learning rates for (A) PascalVOC2012 dataset and (B) ADE20K dataset }
\label{fig:7} 

\end{center}
\end{figure}

A comparative study of our model with some State-Of-The-Art models is given by Table 3 and Table 4 and it is very much evident that proposed MsAUNet framework outperforms others by achieving \textbf{79.88\%} and \textbf{44.88\%} mIoU in PascalVOC2012 and ADE20K datasets respectively. In addition a Piecewise deep learning model, proposed by Lin et al.[33], also brings good mIoU of 75.3\% in PascalVOC2012 dataset. Apart from that, Liu et al [34] proposes DPN model for scene segmentation which also brings promising mIoU of 74.1\% in PascalVOC2012 dataset. In ADE20K dataset EncNet(Zhang et al. [38]) and PSPNet(Zhao et al [37]) achieves good mIoU scores of 41.11\% and 43.29\% with Res101 and Res50 as the backbone architectures. Overall we can conclude that MsAUNet achieves outperforming results while compared with aforementioned State-Of-The-Art models on standard datasets like PascalVOC2012 and ADE20K. 

\begin{table*}[]
\caption{Comparison Table of PascalVOC2012}
\resizebox{\textwidth}{!}{\begin{tabular}{|l|l|l|l|l|l|}
\hline
Methods         & Base Net    & Pixel Accuracy & mIoU    & Frequency Weighted   IoU & Dice Score (F1 Score) \\ \hline
FCN [30]             &             & 88.21\%        & 62.2\%  & 75.34\%                  & 69.35\%               \\ \hline
DeepLav2 [31]          &             & 91.32\%        & 71.6\%  & 79.44\%                  & 68.59\%               \\ \hline
GCRF [32]            &             & 90.02\%        & 73.2\%  & 74.68\%                  & 75.66\%               \\ \hline
Piecewise [33]       &             & 94.12\%        & 75.3\%  & 73.49\%                  & 74.37\%               \\ \hline
DPN [34]             &             & 92.44\%        & 74.1\%  & 77.28\%                  & 76.13\%               \\ \hline
Proposed MsAUNet & DenseNet169 & \textbf{95.96\%}        & \textbf{79.88\%} & \textbf{80.79\% }                 & \textbf{79.01\%}              \\ \hline
\end{tabular}}

\caption{Comparison Table ADE20K}
\resizebox{\textwidth}{!}{\begin{tabular}{|l|l|l|l|l|l|}
\hline
Methods         & Base Net    & Pixel Accuracy & mIoU    & Frequency Weighted   IoU & Dice Score (F1 Score) \\ \hline
FCN [30]             &             & 71.32\%        & 29.39\% & 40.49\%                  & 36.98\%               \\ \hline
CascadeNet [35]       &             & 74.52\%        & 34.90\% & 47.33\%                  & 38.57\%               \\ \hline
SegNet [36]           &             & 71.00\%        & 21.64\% & 34.11\%                  & 35.78\%               \\ \hline
PSPNet [37]           & Res101      & 81.39\%        & 43.29\% & 52.69\%                  & 48.13\%               \\ \hline
EncNet [38]           & Res50       & 79.73\%        & 41.11\% & 52.11\%                  & 45.33\%               \\ \hline
Proposed MsAUNet & DenseNet169 & \textbf{81.41\%}        & \textbf{44.88\%} & \textbf{54.48\%}                  & \textbf{51.01\%}               \\ \hline
\end{tabular}}
\end{table*}

In addition to mIoU, our model shows its better performance than all other popularly used models mentioned above, in the essence of Pixel Accuracy, Frequency Weighted IoU and Dice Accuracy also. Table 4 and Table 5 shows the outperforming results of MsAUNet with good marginal difference. 

Some visual results of our model with DenseNet169 as the backbone architecture on ADE20K dataset is given by Fig.4.

\section{Conclusion and Future Work}
Object detection using various scene segmentation techniques has always been a research area of great interest for computer vision scientists. Such segmentation works have been used in various domains of computer vision such as object detection with the aid of scene segmentation, bio-medical image segmentation and so on. Various models of object detection like one-shot learning models, such as SSD300, YOLOv3 has already been introduced so far. But these models have high degree of architectural complexity with relatively costly annotation datasets. Compare to that, scene segmentation models are relatively easy to build and achieves promising results with less costly frameworks. Hereby we conclude our work with some future scopes in the field of computer vision, with the aid of various segmentation techniques. 
\begin{itemize}
    \item MsAUNet can also be used in the field of bio-medical image processing for cell segmentation tasks.
    \item MsAUNet with DenseNet264 as the encoder model, might assure good performance scores with large datasets of object detection such as COCO dataset. 
    \item Proposed MsAUNet can be implemented in different domains of computer vision, as mentioned above. With different backbone architectures significant changes can be made at different fields of computer visions.  
    \item In addition, different combinations of optimizers and learning rates along with different hybrid loss functions can significantly improve the performance of MsAUNet for different datasets from different fields. 
\end{itemize}

\end{document}